# Close Clustering Based Automated Color Image Annotation

Ankit Garg          Rahul Dwivedi          Dr. Krishna Asawa

*Jaypee Institute of Information Technology (JIIT) University, India*
{*ankitgarg20, rahuldwivedi.jiit*}@gmail.com     *krishna.asawa@jiit.ac.in*

***Abstract*—** *Most image-search approaches today are based on the text based tags associated with the images which are mostly human generated and are subject to various kinds of errors. The results of a query to the image database thus can often be misleading and may not satisfy the requirements of the user. In this work we propose our approach to automate this tagging process of images, where image results generated can be fine filtered based on a probabilistic tagging mechanism. We implement a tool which helps to automate the tagging process by maintaining a training database, wherein the system is trained to identify certain set of input images, the results generated from which are used to create a probabilistic tagging mechanism. Given a certain set of segments in an image it calculates the probability of presence of particular keywords. This probability table is further used to generate the candidate tags for input images.*

***Index Terms*—** Close clustering, image annotation, image search.

## I. INTRODUCTION

With the advent of internet and its easy accessibility all over the world it has become the largest source of images. Getting hold of any kind of image is not that difficult, but the key problem is to get the right one. Most image search portals (such as Google, Flicker and Yahoo etc) mostly rely on human participants to tag images which are sometimes not only time consuming but also error prone. With the increasing need for availability of results in short period of time and served in an efficient manner, manual tagging has proved to be quite an expensive task. In case of manual textual tagging, many images are left untagged and are not shown in the results of the query posted by the user, which may lead to loss of important results. The difficulty arises from the fact that several people usually use different tags for the same document. A user capturing a scene of a "rose flower" with a canon camera at some park in Delhi may tag it as "Canon, flower, Delhi", but its semantic description in terms of tags may not be sufficient enough for it to be searched through the image database.

With more and more search engines coming up and trying to be more efficient than others, it has become extremely important to develop a tool that increases the efficiency of information retrieval, using least human intervention.

Auto-tagging of images could be one possible approach of applying system generated text descriptions to an image so that they can be easily categorised and searched when queried upon by a user. However, the main hurdle in a system assisted tagging is classification of objects. There have been multiple approaches based on image segmentation and in our work we implement segmentation using close clustering algorithm, the name so derived because of its ability to cluster data depending upon its closeness to the already present cluster. We further provide a probabilistic approach for annotation of the images based on the segments from a training dataset.

## II. PROBLEM STATEMENT

This work is aimed at designing an application that could auto-tag a collection of images. The need for auto-tagging arises from the fact that manual tagging is both expensive in terms of effort and results generated. Manual tagging is guided by human perception which may vary from human to human (a single image can have different sets of overlapping tags), and could be error prone. We attempt to mitigate the human intervention in this tagging process, to an extent where a system can utilize an initial training dataset (which is created by a set of users) and use that for automatic image tag generation.

## III. PRIOR WORK

There has been considerable work done in the field of intelligent image tagging. Some initial efforts have recently been devoted to automatically annotating pictures, leveraging decades of research in multiple areas like computer vision, image understanding, image processing, and statistical learning. Also there exists significant number of papers which have applied generative modelling, statistical boosting, visual templates, multiple instance learning, active learning, feedback learning and manifold learning, to problems dealing with image classification, annotation, and retrieval.

Normally the tagging process can be classified into two types, namely descriptive tagging, and people tagging. Descriptive tagging is used to identify various objects in an image and even human emotions. People tagging involves associating tags with people. *Tagcow* accomplices the tagging task by hiring a third party worker and when the task is complete it offers its user the new tagged image for a certain monetary cost. But this still involves human intervention and can be erroneous.

However one of a recent work by Jia.Lee et al [8] overcomes this problem to a considerable extent. Developed by Penn State researchers in 2005, it was made public in October 2005. At present it has a vocabulary of around 330 words. Given an image or the image URL, this application presents 15 most probable annotation or tags for the image. The system has less vocabulary and its accuracy is also low.

Another significant work in this field is the image search engine known as *Behold*. It is based on statistical models for 'automatic image annotation'. The search engine based on [10] is capable of recognising a number of visual concepts in pictures by the use of visual filters. You can search images that look like one of these concepts.

## IV. PROPOSED SOLUTION

In our work, we base our solution for automated image-tagging on a probabilistic approach for annotation of the images based on the segments and a training dataset. We implement our approach in 5 sequential steps. Steps I-IV is used to create the initial database. An image that needs to be tagged undergoes steps V – IX.

i. Segmentation – A necessary step to segregate the objects from an image.
ii. Blob tokens generation – Objects retrieved from all the images participating in the training database, need to be grouped together so that similar objects are kept together in a form of cluster.
iii. Training database – Blob tokens need to be stored, for formation of probability table.
iv. Probability table – Associating of keywords with the blobs.
v. Segment the input image.
vi. Find the blob representation of the input image.
vii. Use the table generated to predict the tags for an input image.
viii. Generation of tags.
ix. Choosing the best possible tags.

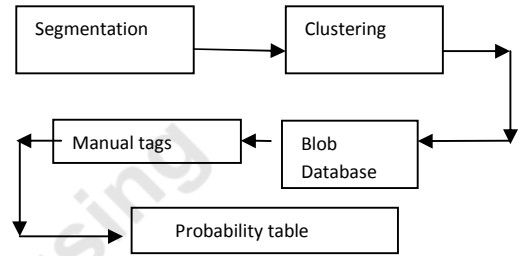

Fig. 1

Fig. 1 represents the sequential steps in creating the training database. We detail the steps in entailing sections.

### A. Segmentation

Tagging requires each object to be classified separately, and hence it is important to perform segmentation of the image. The whole success of the results of the tool lies on the process of segmentation, better the objects are identified well would be their classification and finally better be the tagging of images. There are various algorithms available for performing segmentation, but depending upon the need of the tool they may suit to varying degree. We tried upon some of the algorithms already available such as the one based on discrete wavelet transform (DWT) [9] followed by mean shift filtering but the results were not encouraging enough, the segmentation obtained was not clear enough to carry on further work. The segmentation algorithm we have applied is completely new and uses clustering technique that is totally different from the available clustering technique.

We had a close look on traditional clustering algorithms, specifically the K- means algorithm and other less known algorithms such as simple link, complete link, and graph clustering [7]. The basic drawback which deterred us from using K-means algorithm was it required selection of initial partition, we needed a technique which would automatically give us the required number of clusters, without being specified the original number of clusters. The database for images being used is COREL database. All the images chosen for the training set undergo segmentation one by one. Image segmentation can be applied in a series of steps.

*Calculating RGB components* – For each pixel of the image its RGB component is calculated, where R, G, B respectively stand for red, green, blue colour components in an image, (r ,g, b, X,Y,Z) are temporary values in the process of conversion.

$$[XYZ] = [rgb][M] \tag{1}$$

$$r = \begin{cases} \frac{R}{12.92} & R \leq 0.4045 \\ \left(\frac{R+0.055}{1.055}\right)^{24} & R > 0.4045 \end{cases} \tag{2}$$

$$g = \begin{cases} \frac{G}{12.92} & G \leq 0.4045 \\ \left(\frac{G+0.055}{1.055}\right)^{24} & G > 0.4045 \end{cases} \tag{3}$$

$$b = \begin{cases} \frac{B}{12.92} & B \leq 0.4045 \\ \left(\frac{B+0.055}{1.055}\right)^{24} & B > 0.4045 \end{cases} \tag{4}$$

$$M = \begin{pmatrix} 0.412424 & 0.212656 & 0.0193324 \\ 0.357579 & 0.715158 & 0.1191930 \\ 0.180464 & 0.0721856 & 0.950444 \end{pmatrix} \tag{5}$$

***RGB to Luv conversion*** – L-luminescence, U-saturation, V - hue angle components are considered to be more appropriate than RGB components are due to its coherence with human eye perception Conversion requires the use of intermediate references u', v'.

$$L = \begin{cases} 116\sqrt[3]{y_r} - 16 & y_r > \varepsilon \\ k y_r & y_r \leq \varepsilon \end{cases} \tag{6}$$

$$u = 13L(u' - u_r') \tag{7}$$
$$v = 13L(v' - v_r') \tag{8}$$
$$y_r = \frac{y}{y_r} \tag{9}$$
$$u' = \frac{4X}{X + 15Y + 3Z} \tag{10}$$
$$v' = \frac{9Y}{X + 15Y + 3Z} \tag{11}$$
$$u_r' = \frac{4X_r}{X_r + 15Y_r + 3Z_r} \tag{12}$$
$$v_r' = \frac{9Y_r}{X_r + 15Y_r + 3Z_r} \tag{13}$$

We further use these two constants given by CIE standard :

$\varepsilon = 0.0085$     Actual CIE Standard     (14)

$\varepsilon = \frac{216}{24389}$     Intent of the CIE Standard     (15)

$k = 903.3$     Actual CIE Standard     (16)

$k = \frac{\frac{216}{24389}}{27}$     Intent of the CIE Standard     (17)

We were concerned only with the L component of each pixel of the image. Clustering of each and every pixel based upon its L value provides us with different segments. Initially each pixel is considered as a single and separate cluster . Each pixel's L value is compared with every other pixel, and those pixels whose value lie in a certain range above and below the L value of the pixel being compared is allocated to that pixel's cluster.
Where Range = (max ($L_{value}$)- min($L_{value}$))/$N_c$
and Nc = number of clusters.

The centroid value of the cluster is the mean value of the original cluster and the newly added cluster. The algorithm is an iterative process, a new range is defined after avery single iteration until all pixels have been assigned to proper clusters.

The steps are performed on every image of the training set of images, so as to generate segments in each of the image.

**B. Blob tokens generation**
Once we have segmented all our images, the next step is to group similar segments obtained from these images, hence we apply the clustering technique again on the available segments. This provides us with new set of clusters , where each cluster contains a specific set of regions extracted from different images of the database. Each new cluster is now called as a blob token. Blob tokens are set of segments that denote similar segments and have a centroid value that acts as a signature for that blob.

**C. Training database creation**
After we prepare our blob tokens, we generate a list of each segment of every image with us. We store the details of each image segment for the purpose of generation of probability table in a XML file. As shown below this XML schema comprises of nodes consisting of the blob number, segment number, image number , value of the blob centroid and the pixel stored under each blob in a hierarchial form with a single root node.

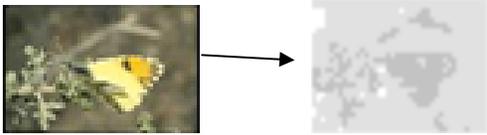

Fig .2 Close Clustering based Image Segmentation

```
<Root node>
        <child node 1>
        <text node>text</text node>
        </child node>
        <child node 2>
        <text node>text</text node>
        < /child node>
</Root node>
```

Depending upon the requirement the node structure can be made more deeper. Each image in the database is now represented as a set of blobs from the blob vocabulary. Further, the images in the training set are manually annotated. Keywords are provided for each image; these words represent each image as whole and are not assigned to an individual segment/region. Thus each image has a dual representation in terms of a set of blob tokens and word tokens ($b_1…b_m$ represent the blob tokens and $w_1…w_n$ represent the words manually assigned to images)

$J = \{b_1 . . . b_m; w_1 . . . w_n\}$.

The next step is to link the blob tokens with the word tokens, which is accomplished using the probability table.

### D. Probability table generation

After the creation of the training database, our key problem is reduced to that of using the training data set to construct a probability table linking blob tokens with word tokens. This table is the conditional probability of a word token given a blob token. Each word is predicted with some probability by each blob, meaning that we have a mixture model for each word.

We estimate the probability $P(w|I)$ for every word $w$ in the vocabulary. The probability of drawing the word $w$ is best approximated by the conditional probability of observing $w$ given that we previously observed $b_1 . . . b_m$ as a random sample from the same distribution:

$$P(w|I) \approx P(w|b_1...b_m)$$

We use the training set database previously created to estimate the joint probability of observing the word w and the blobs $b_1..........b_m$ in the same image.

A matrix is generated $M_{N*(W+B)}$ where
N= number of images
W=number of keywords
B= number of blobs

Each segment of every image is allocated to one or the other keyword and from the XML file we can find out which segment is part of which blob and thus we can have information about the blob numbers constituting an image. The matrix forming keywords is called $M_W$ whereas the one forming blobs is called as $M_B$. A new matrix is formed upon calculation of $M^T_W * M_B$ which on normalisation gives probability $p(W_i|B_j)$ where, $p(W_i|B_j)=$ conditional probibility of keyword $W_i$ given blob $B_j$. Thus each blob is now associated with certain keywords.

### E. Generation of blob representation

This step is mainly used to predict most probable annotations for a given un-annotated image *I*. We segment the image into regions and compute the region features. After comparing each segment to the clustered regions, we assign the blob that is closest to it in the previously obtained cluster space in the training database.

We consider the blob representation of the image in the form $I = \{b_1 . . . b_m\}$, where I represents the image comprising of blob tokens $b_1...b_m$

Here, we need to automatically select a set of words *{$w_1 ...w_n$} that* accurately reflects the content of the image. We derive these set of words from the previously annotated images in the training set.

### F. Choosing best possible tags

Since the number of tags selected depends upon the number of blobs detected. It may happen that we are available with too many tags thus decreasing its accuracy; hence our application selects the best possible 15 keywords from all the keywords detected.

This is done by arranging all the keywords generated above in order of their probability values and then choosing among them the best 15.

## V. EVALUATION OF OUR APPROACH

For testing our system we took a database of 100 images, consisting of 5 groups of images, each consisting of 20 images. Our images for training set were taken from the COREL image database. The tests were performed on 78 images belonging to the same dataset of images. The work was tested on different training databases, with databases varying with respect to the number of images being used.

Out of the 78 images, 79 % showed accuracy greater than 50 %. 91% percent images showed at least one correct annotation. The results showed that higher the number of images in the training database higher was the accuracy of the results generated. This arises from the fact that higher the number of images, higher is the number of segments formed; this leads to better clusters formed. Better clusters lead to higher probability of associating blob tokens with keywords.

Table.1 Number of images Vs Percentage Accuracy

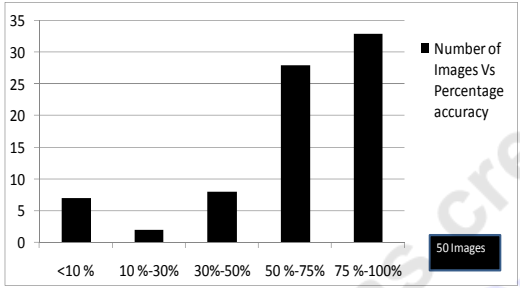

Table.2 Percentage of Images Vs Percentage Accuracy

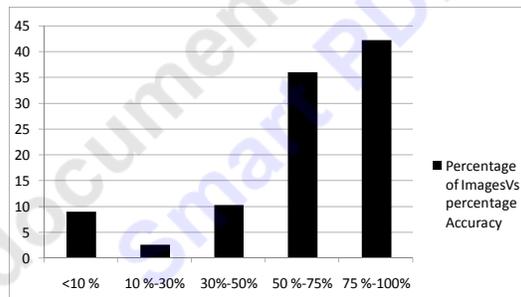

## VI. CONCLUSIONS AND FUTURE WORK

Our process undertaken to perform automation is highly database centric and would definitely improve with large set of images. Further, since the process is dependent on segmentation, the efficiency of our system improves as we use a better segmentation algorithm. The algorithm discussed here is better than DWT method since the clustering algorithm designed here is such that it chooses the closest pixels to a group before performing clustering. Also, during the formation of the matrix M, each segment is allocated to a keyword. This is quite tedious job as it requires a lot of time and concentration, because for any dataset if some of the keywords are allocated to a wrong segment, then it can have a large scale affect on the finally generated tags, thus saving our clusters in a file solves that problem to a great extent.